\documentclass[10pt,twocolumn,letterpaper]{article}

\usepackage[pagenumbers]{cvpr} 

\usepackage{graphicx}
\usepackage{amsmath}
\usepackage{amssymb}
\usepackage{booktabs}

\usepackage{amsfonts}       
\usepackage{nicefrac}       
\usepackage{microtype}      
\usepackage{xcolor}         
\usepackage{graphicx}
\usepackage{diagbox}
\usepackage{multicol}
\usepackage{enumerate}
\usepackage{times}
\usepackage{epsfig}
\usepackage{graphicx}
\usepackage{amsmath}
\usepackage{amssymb}
\usepackage{threeparttable}
\usepackage{enumitem}
\usepackage{multirow}
\usepackage{color}
\usepackage{array}
\usepackage{bbding}
\usepackage{makecell}
\usepackage[pagebackref,breaklinks,colorlinks]{hyperref}

\usepackage[capitalize]{cleveref}
\crefname{section}{Sec.}{Secs.}
\Crefname{section}{Section}{Sections}
\Crefname{table}{Table}{Tables}
\crefname{table}{Tab.}{Tabs.}

\begin{document}

\title{Unsupervised Cumulative Domain Adaptation for Foggy Scene Optical Flow}

\author{Hanyu Zhou\textsuperscript{\rm 1}, Yi Chang\textsuperscript{\rm 1}\thanks{Corresponding author.}, Wending Yan\textsuperscript{\rm 2}, Luxin Yan\textsuperscript{\rm 1}\\
  \textsuperscript{\rm 1} National Key Laboratory of Science and Technology on Multispectral Information Processing,\\School of Artificial Intelligence and Automation, Huazhong University of Science and Technology\\
  \textsuperscript{\rm 2} Huawei International Co. Ltd.\\
  {\tt\small {\{hyzhou, yichang, yanluxin\}}@hust.edu.cn, yan.wending@huawei.com}
 }

\maketitle


\begin{abstract}

Optical flow has achieved great success under clean scenes, but suffers from restricted performance under foggy scenes. To bridge the clean-to-foggy domain gap, the existing methods typically adopt the domain adaptation to transfer the motion knowledge from clean to synthetic foggy domain. However, these methods unexpectedly neglect the synthetic-to-real domain gap, and thus are erroneous when applied to real-world scenes. To handle the practical optical flow under real foggy scenes, in this work,  we propose a novel unsupervised cumulative domain adaptation optical flow (UCDA-Flow) framework: depth-association motion adaptation and correlation-alignment motion adaptation. Specifically, we discover that depth is a key ingredient to influence the optical flow: the deeper depth, the inferior optical flow, which motivates us to design a depth-association motion adaptation module to bridge the clean-to-foggy domain gap. Moreover, we figure out that the cost volume correlation shares similar distribution of the synthetic and real foggy images, which enlightens us to devise a correlation-alignment motion adaptation module to distill motion knowledge of the synthetic foggy domain to the real foggy domain. Note that synthetic fog is designed as the intermediate domain. Under this unified framework, the proposed cumulative adaptation progressively transfers knowledge from clean scenes to real foggy scenes. Extensive experiments have been performed to verify the superiority of the proposed method.

\end{abstract}

 \begin{figure}
   \setlength{\abovecaptionskip}{5pt}
   \setlength{\belowcaptionskip}{-5pt}
   \centering
    \includegraphics[width=1.0\linewidth]{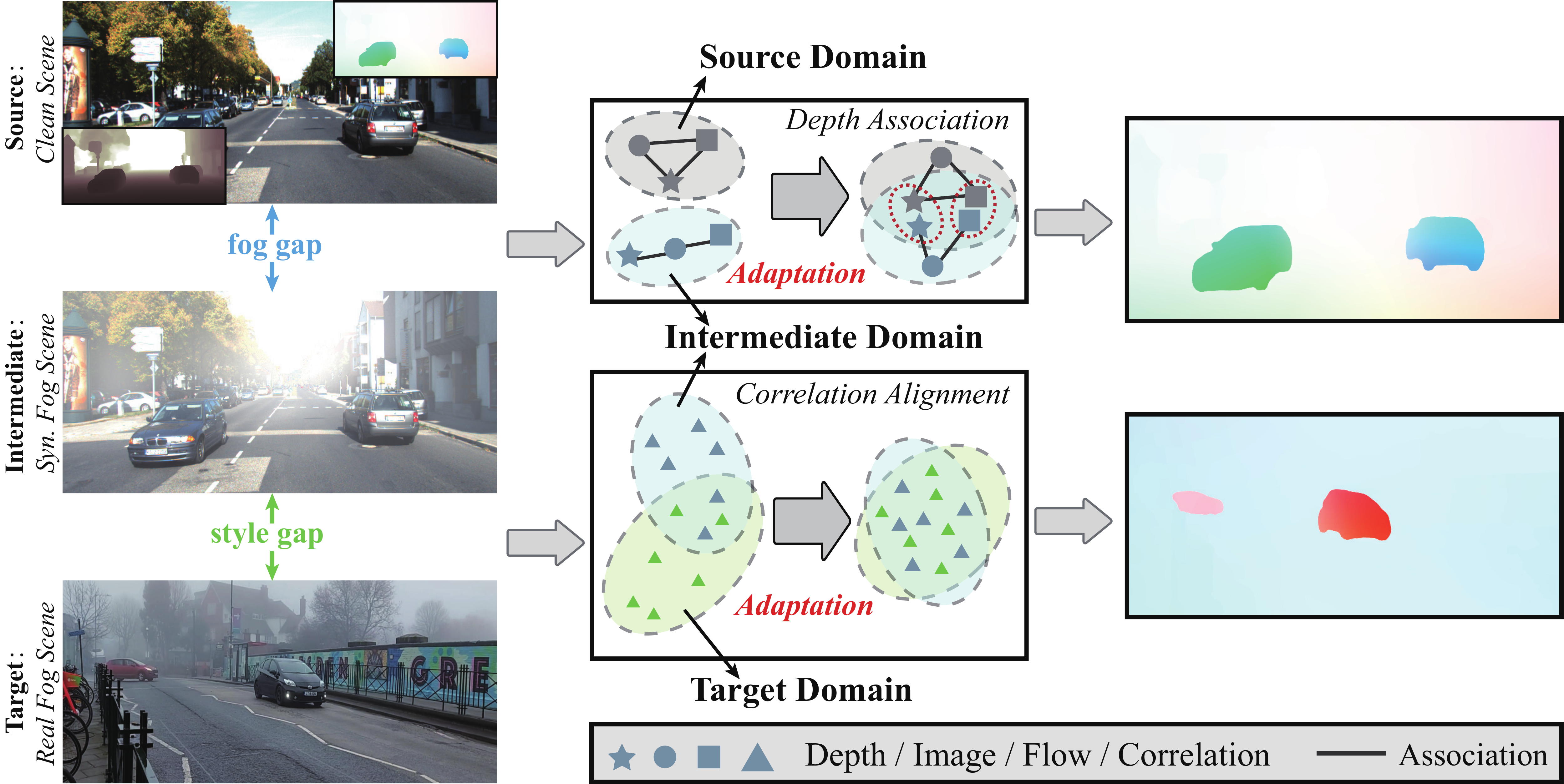}
    \caption{Illustration of the main idea. We propose to transfer motion knowledge from the source domain (clean scene) to the target domain (real foggy scene) through two-stage adaptation. We design the synthetic foggy scene as the intermediate domain. As for the clean-to-foggy domain gap (fog), we transfer motion knowledge from the source domain to the intermediate domain via depth association. As for the synthetic-to-real domain gap (style), we distill motion knowledge of the intermediate domain to the target domain by aligning the correlation of both the domains.
    }
    \label{Main_Idea}

 \end{figure}

\section{Introduction}
\label{sec:intro}
Optical flow has made great progress under clean scenes, but may suffer from restricted performance under foggy scenes \cite{li2018robust}. The main reason is that fog weakens scene contrast, breaking the brightness and gradient constancy assumptions, which most optical flow methods rely on.

To alleviate this, researchers start from the perspective of domain adaptation, which mainly seeks the degradation-invariant features to transfer the motion knowledge from the clean scene to the adverse weather scene \cite{li2018robust, li2019rainflow, yan2020optical, li2021gyroflow}. For example, Li \cite{li2018robust, li2019rainflow} attempted to learn degradation-invariant features to enhance optical flow under rainy scenes in a supervised manner.
Yan \emph{et al.} \cite{yan2020optical} proposed a semi-supervised framework for optical flow under dense foggy scenes, which relies on the motion-invariant assumption between the paired clean and synthetic foggy images. These pioneer works have made a good attempt to handle the clean-to-foggy domain gap with synthetic degraded images through one-stage domain adaptation.
However, they lack the constraints to guide the network to learn the motion pattern of real foggy domain, and fail for real foggy scenes. In other words, they have unexpectedly neglected the synthetic-to-real domain gap, thus limiting their performances on real-world foggy scenes.
In this work, our goal is to progressively handle the two domain gaps: the clean-to-foggy gap and the synthetic-to-real gap in a cumulative domain adaptation framework in Fig. \ref{Main_Idea}.

As for the clean-to-foggy gap, we discover that depth is a key ingredient to influence the optical flow: the deeper the depth, the inferior the optical flow. This observation inspires us to explore the usage of depth as the key to bridging the clean-to-foggy gap (seeing the fog gap in Fig. \ref{Main_Idea}). On one hand, depth physically associates the clean image with the foggy image through atmospheric scattering model \cite{narasimhan2002vision}; on the other hand, there exists a natural 2D-3D geometry projection relationship between depth and optical flow, which is used as a constraint to transfer motion knowledge from the clean domain to the synthetic foggy domain.

As for the synthetic-to-real gap, we figure out that cost volume correlation shares similar distribution of synthetic and real foggy images.
Cost volume stores correlation value, which can physically measure the similarity between adjacent frames, regardless of synthetic and real foggy images. Therefore, cost volume benefits to bridging the synthetic-to-real domain gap (seeing the style gap in Fig. \ref{Main_Idea}). We align the correlation distributions to distill motion knowledge of the synthetic foggy domain to the real foggy domain.

In this work, we propose a novel unsupervised cumulative domain adaptation optical flow (UCDA-Flow) framework for real foggy scene, including depth-association motion adaptation (DAMA) and correlation-alignment motion adaptation (CAMA).
Specifically, in DAMA stage, we first estimate optical flow, ego-motion and depth with clean stereo images, and then project depth into optical flow space with 2D-3D geometry formula between ego-motion and scene-motion to enhance rigid motion. To bridge the clean-to-foggy gap, we utilize atmospheric scattering model \cite{narasimhan2002vision} to synthesize the corresponding foggy images, and then transfer motion knowledge from the clean domain to the synthetic foggy domain. In CAMA stage, to bridge the synthetic-to-real domain gap, we transform the synthetic and real foggy images to the cost volume space, in which we align the correlation distribution to distill the motion knowledge of the synthetic foggy domain to the real foggy domain. The proposed cumulative domain adaptation framework could progressively transfer motion knowledge from clean domain to real foggy domain via depth association and correlation alignment. Overall, our main contributions are summarized as follows:

\begin{itemize}[leftmargin=10pt]
\item We propose an unsupervised cumulative domain adaptation framework for optical flow under real foggy scene, consisting of depth-association motion adaptation and correlation-alignment motion adaptation. The proposed method can transfer motion knowledge from clean domain to real foggy domain through two-stage adaptation.

\setlength{\itemsep}{-2pt}
\item We reveal that foggy scene optical flow deteriorates with depth. The geometry relationship between depth and optical flow motivates us to design a depth-association motion adaptation to bridge the clean-to-foggy domain gap.

\setlength{\itemsep}{-2pt}
 \item We illustrate that cost volume correlation distribution of the synthetic and real foggy images is consistent. This prior benefits to close the synthetic-to-real domain gap through correlation-alignment motion adaptation.

\end{itemize}

\section{Related Work}
\label{sec:related_work}
\noindent
\textbf{Optical Flow.} Optical flow is the task of estimating per-pixel motion between video frames. Traditional methods \cite{sun2010secrets} often formulate optical flow as an energy minimization problem.
In recent years, the learning-based optical flow approaches \cite{dosovitskiy2015flownet, jason2016back, ilg2017flownet, ranjan2017optical, lai2017semi, ren2017unsupervised, sun2018pwc, hui2018liteflownet, liu2020learning, luo2021upflow, chi2021feature, zhang2021separable, jiang2021learning, aleotti2021learning, sun2021autoflow, huang2022flowformer} have been proposed to improve the feature representation.
PWC-Net \cite{sun2018pwc} applied warp and cost volume to physically estimate optical flow in a coarse-to-fine pyramid.
RAFT \cite{teed2020raft} was an important development of PWC-Net, which replaced the pyramid architecture with GRU \cite{cho2014properties} and constructed 4D cost volume for all pairs of pixels.
To improve the motion feature representation, GMA \cite{jiang2021transformer} incorporated transformer into optical flow estimation and achieved better performance than RAFT.
Furthermore, to relieve the dependency on synthetic datasets, the authors \cite{jason2016back, ren2017unsupervised, yang2019volumetric, meister2018unflow, liu2019ddflow, zhao2020maskflownet, jonschkowski2020matters} proposed the unsupervised CNN optical flow methods with photometric loss or data distillation loss.
Although they have achieved satisfactory results in clean scenes, they would suffer from degradation under foggy scenes. In this work, we propose an unsupervised optical flow framework for real foggy scenes.

\noindent
\textbf{Optical Flow under Adverse Weather.}
The robust optical flow estimation has been extensively studied for various adverse weather, such as rain \cite{li2019rainflow}, fog \cite{yan2020optical}.
An intuitive solution to this challenging task is to perform the image deraining \cite{fu2017clearing, yang2017deep, zhang2018density, liu2021unpaired, yan2021self} or defogging \cite{ren2018gated, liu2019griddehazenet, qin2020ffa, shao2020domain, wu2021contrastive} with subsequent optical flow estimation.
However, existing derain/defog methods are not designed for optical flow and the possible over-smoothness or residual artifacts would contribute negative to optical flow.
To bridge the clean-to-degraded gap, the authors \cite{li2018robust, li2019rainflow, yan2020optical, li2021gyroflow} have attempted to design the domain-invariant features to transfer motion knowledge from clean domain to synthetic degraded domain through one-stage adaptation. For example, Li \emph{et al}. \cite{li2018robust, li2019rainflow} attempted to design rain-invariant features in a unified framework for robust optical flow under rainy scenes with synthetic degraded images.
Yan \emph{et al}. \cite{yan2020optical} estimated optical flow under dense foggy scenes via optical flow consistency.
Li \emph{et al}. \cite{li2021gyroflow} resorted to auxiliary gyroscope information which is robust to degradation for adverse weather optical flow.
To further close the synthetic-to-real domain gap, we propose a two-stage cumulative domain adaptation framework for optical flow under real foggy scenes, which can bridge the clean-to-foggy and synthetic-to-real domain gaps.
%


 \begin{figure*}
   \setlength{\abovecaptionskip}{5pt}
   \setlength{\belowcaptionskip}{-10pt}
   \centering
    \includegraphics[width=0.99\linewidth]{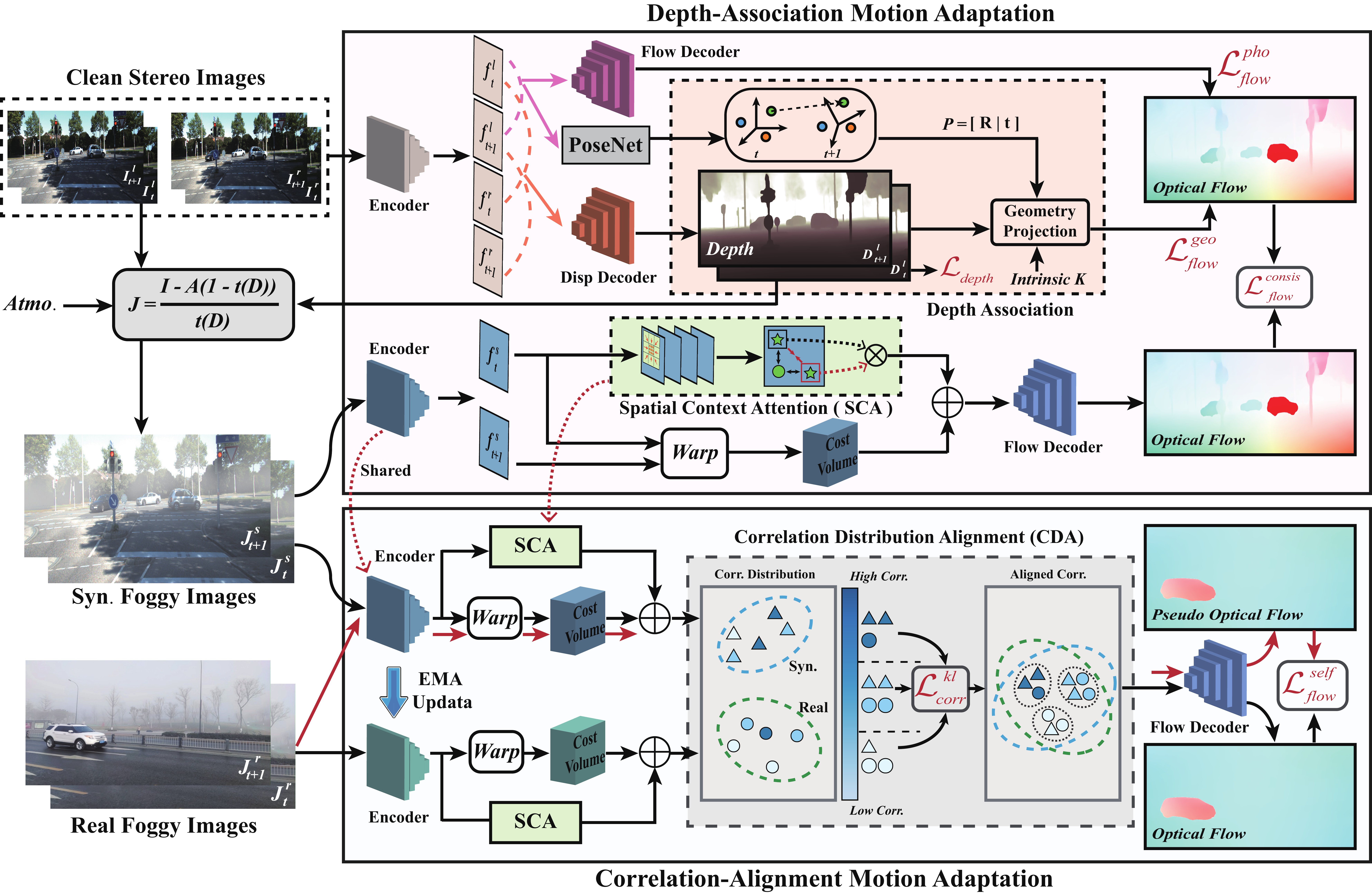}
    \caption{The architecture of the UCDA-Flow mainly contains depth-association motion adaptation (DAMA) and correlation-alignment motion adaptation (CAMA). The goal of DAMA stage is to bridge the clean-to-foggy domain gap, in which we associate the depth with optical flow via geometry projection and synthetic foggy images with atmospheric scattering model, and transfer motion knowledge from the clean domain to the synthetic foggy domain. In CAMA stage, to further close the synthetic-to-real domain gap, we align the correlation distributions of synthetic and real foggy images to distill motion knowledge of the synthetic foggy domain to the real foggy domain.}
    \label{Framework}
 \end{figure*}

\section{Unsupervised Cumulative Adaptation}
\label{sec:method}
\subsection{Overall Framework}
The goal of this work is to estimate optical flow under real foggy scenes. Most existing adverse weather optical flow methods mainly adopt the one-stage adaptation to transfer motion knowledge from clean domain to synthetic adverse weather domain. However, due to the synthetic-to-real domain gap that these methods neglect, they cannot generalize well for real degraded scenes. In this work, we illustrate that foggy scene optical flow deteriorates with depth, which can bridge the clean-to-foggy domain gap. Moreover, we figure out that cost volume correlation shares the similar distribution of synthetic and real foggy images, benefiting to bridge the synthetic-to-real domain gap. Motivated by these analyses, we propose a novel unsupervised cumulative domain adaptation framework for optical flow under real foggy scenes. As shown in Fig. \ref{Framework}, our framework consists of two main modules: depth-association motion adaptation (DAMA) and correlation-alignment motion adaptation (CAMA). The DAMA associates depth with optical flow via geometry projection, renders synthetic foggy images with atmospheric scattering model, and transfers motion knowledge from clean domain to synthetic foggy domain. The CAMA aligns the correlation distribution of synthetic and real foggy images to distill motion knowledge of synthetic foggy domain to real foggy domain. Under this unified framework, the proposed framework could progressively transfer motion knowledge from clean domain to real foggy domain.

\subsection{Depth-Association Motion Adaptation}
The previous methods \cite{li2019rainflow, yan2020optical} have attempted to directly transfer motion knowledge from clean domain to synthetic degraded domain. However, different from rain and snow, fog is a non-uniform degradation related to depth. This makes us naturally consider whether the optical flow affected by fog could be related to depth or not.

\begin{figure}
  \setlength{\abovecaptionskip}{5pt}
  \setlength{\belowcaptionskip}{-5pt}
  \centering
   \includegraphics[width=1.0\linewidth]{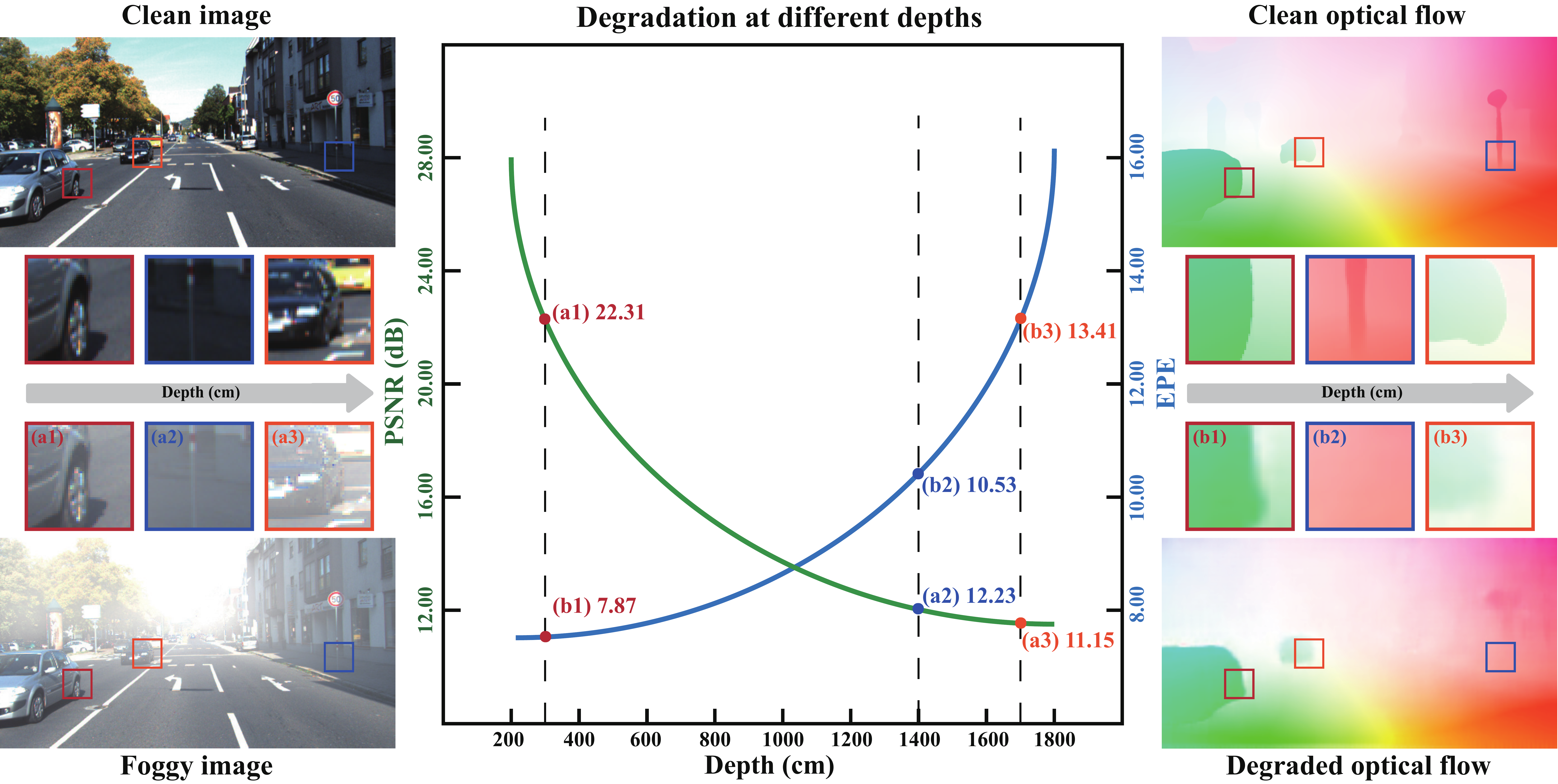}
   \caption{Analysis of fog degradation at different depths. The deeper the depth, the more severe the image and the optical flow. Depth is the key to bridging the clean-to-foggy domain gap.}
   \label{Depth_Experment}
\end{figure}

To illustrate this, we conduct an analysis experiment on the influence of fog on the image and optical flow along different depths in Fig. \ref{Depth_Experment}.
We take clean KITTI2015 \cite{menze2015object} and synthetic Fog-KITTI2015 as the experimental datasets. Compared to the corresponding clean images, we count the PSNR and the optical flow EPE of the foggy images at different depths.
As the depth value becomes larger, the lower the PSNR, the higher the optical flow EPE, which means that the degradation of the image and optical flow aggravates with the larger depth. Moreover, we visualize the images (Fig. \ref{Depth_Experment} (a1)-(a3)) and optical flows (Fig. \ref{Depth_Experment} (b1)-(b3)) at three depths. We can observe that the contrast of images and boundaries of optical flows become more blurry with the increasing of the depth.
This inspires us that depth is the key to bridging the clean-to-foggy domain gap.
On one hand, depth is associated with fog through atmospheric scattering model \cite{narasimhan2002vision}, which bridges the clean-to-foggy domain gap; on the other hand, depth could be used to refine optical flow of clean domain via strict geometry projection, and serve as a constraint to transfer motion knowledge from the clean domain to the synthetic foggy domain.
Therefore, we propose a depth-association motion adaptation module to transfer motion knowledge between both the domains.

\noindent
\textbf{Depth Association.} Given consecutive stereo images ${[\textbf{\emph{I}}_{t}^{l}, \textbf{\emph{I}}_{t+1}^{l}, \textbf{\emph{I}}_{t}^{r}, \textbf{\emph{I}}_{t+1}^{r}]}$, we first take RAFT \cite{teed2020raft} to estimate optical flow ${\textbf{{\emph{F}}}}$ with optical flow photometric loss \cite{jason2016back} in an unsupervised manner as follow,
\begin{equation}
  \setlength\abovedisplayskip{3pt}
  \setlength\belowdisplayskip{3pt}
\begin{aligned}
\resizebox{0.10\hsize}{!}{$\mathcal{L}^{pho}_{flow}$} &=  \resizebox{0.75\hsize}{!}{$\sum\nolimits{\psi(\textbf{\emph{I}}_{t}^{l} - warp(\textbf{\emph{I}}_{t+1}^{l}))}\odot(1-O_f)/\sum\nolimits{(1-O_f)}$} \\
  &\resizebox{0.80\hsize}{!}{$+\sum\nolimits{\psi(\textbf{\emph{I}}_{t+1}^{l} - warp(\textbf{\emph{I}}_{t}^{l}))}\odot(1-O_b)/\sum\nolimits{(1-O_b)},$}
  \label{eq:photo_loss}
\end{aligned}
\end{equation}
where \emph{warp} is the warping operator, $\psi$ is a sparse $L_p$ norm ($p = 0.4$). $O_f$ and $O_b$ are the forward and backward occlusion mask by checking forward-backward consistency, and $\odot$ is a matrix element-wise multiplication. Similar to optical flow, stereo depths ${[\textbf{\emph{D}}_{t}^{l}, \textbf{\emph{D}}_{t+1}^{l}]}$ can be obtained by DispNet \cite{xu2020aanet} via photometric loss and smooth loss,
\begin{equation}
  \setlength\abovedisplayskip{3pt}
  \setlength\belowdisplayskip{3pt}
  \begin{aligned}
\resizebox{0.11\hsize}{!}{$\mathcal{L}_{depth}$} &= \resizebox{0.60\hsize}{!}{$\sum\nolimits{\psi(\textbf{\emph{I}}_{t}^{l} - warp(\textbf{\emph{I}}_{t}^{r}))} + |{\triangledown}^2\textbf{\emph{D}}_{t}^{l}|e^{-|{\triangledown}^2\textbf{\emph{I}}_{t}^{l}|}$} \\
& \resizebox{0.75\hsize}{!}{$+ \sum\nolimits{\psi(\textbf{\emph{I}}_{t+1}^{l} - warp(\textbf{\emph{I}}_{t+1}^{r}))} + |{\triangledown}^2\textbf{\emph{D}}_{t+1}^{l}|e^{-|{\triangledown}^2\textbf{\emph{I}}_{t+1}^{l}|}$}.
 \label{eq:depth}
 \end{aligned}
\end{equation}

Here we wish to establish the dense pixel correspondence between the two adjacent frames through depth. Let ${\emph{p}_{t}}$ denotes the 2D homogeneous coordinate of an pixel in frame ${\textbf{\emph{I}}_{t}^{l}}$ and ${\textbf{\emph{K}}}$ denotes the camera intrinsic matrix. We can compute the corresponding point of ${\emph{p}_{t}}$ in frame ${\textbf{\emph{I}}_{t+1}^{l}}$ using the geometry projection equation \cite{zhou2017unsupervised},
\begin{equation}
  \setlength\abovedisplayskip{3pt}
  \setlength\belowdisplayskip{3pt}
\begin{aligned}
  \emph{p}_{t+1} = \textbf{\emph{K}}\textbf{\emph{P}}\textbf{\emph{D}}_{t}^{l}(\emph{p}_{t})\textbf{\emph{K}}^{-1}\emph{p}_{t},
  \label{eq:geometry_motion}
\end{aligned}
\end{equation}
where ${\textbf{\emph{P}}}$ is the relative camera motion estimated by the pre-trained PoseNet \cite{kendall2015posenet}. We can then compute the rigid flow ${\textbf{\emph{F}}_{rigid}}$ at pixel ${\emph{p}_{t}}$ in ${\textbf{\emph{I}}_{t}^{l}}$ by,
${\textbf{\emph{F}}_{rigid}(\emph{p}_{t}) = \emph{p}_{t+1} - \emph{p}_{t}}$.
We further enhance motion in rigid regions with the consistency between the geometrically computed rigid flow and the directly estimated optical flow,
\begin{equation}
  \setlength\abovedisplayskip{3pt}
  \setlength\belowdisplayskip{3pt}
  \begin{aligned}
\resizebox{0.11\hsize}{!}{$\mathcal{L}^{geo}_{flow}$} = \resizebox{0.65\hsize}{!}{$\sum\nolimits{||\textbf{\emph{F}} - \textbf{\emph{F}}_{rigid}}||_{1}\odot(1-V)/\sum\nolimits{(1-V)},$}
 \label{eq:geo_flow}
 \end{aligned}
\end{equation}
where \emph{V} denotes the non-rigid region extracted from stereo clean images by forward-backward consistency check \cite{zou2018df} .

\noindent
\textbf{Motion Knowledge Transfer.} To associate depth with fog, we synthesize the foggy images ${[\textbf{\emph{J}}_{t}^{s}, \textbf{\emph{J}}_{t+1}^{s}]}$ corresponding to the clean images using atmospheric scattering model \cite{narasimhan2002vision},
\begin{equation}\small
  \setlength\abovedisplayskip{3pt}
  \setlength\belowdisplayskip{3pt}
  \begin{aligned}
  \resizebox{0.02\hsize}{!}{${\textbf{\emph{J}}}$} = \resizebox{0.25\hsize}{!}{$\frac{\textbf{\emph{I}} - \textbf{\emph{A}}(1 - \emph{t}(\textbf{\emph{D}})}{\emph{t}(\textbf{\emph{D}})},$}
 \label{eq:atmo_scartter}
 \end{aligned}
\end{equation}
where ${\textbf{\emph{A}}}$ denotes the predefined atmospheric light. ${\emph{t}(\cdot)}$ is a decay function related depth.
We then encode the synthetic foggy images into motion features ${[{f}^{s}_{t}, {f}^{s}_{t+1}]}$, and compute the temporal cost volume ${\emph{cv}_{temp} = ({f}^{s}_{t})^{T} \cdot w({f}_{t+1}^{s})}$, where \emph{T} denotes transpose operator and \emph{w} is the warp operator.
Note that, in order to enable the flow model to have a suitable receptive field for smooth constraint of motion feature, we employ a spatial context attention (\textbf{SCA}) module with a non-local strategy \cite{luo2022learning}.
Specifically, we devise a sliding window with a learnable kernel on the motion feature ${f}^{s}_{t}$ to match the non-local similar feature ${{f}_{sim}}$, and generate ${\emph{k}}$ similar features corresponding to the cropped features from the motion feature ${f}^{s}_{t}$ during sliding searching, as ${[{f}_{sim}^{1}, {f}_{sim}^{2}, ..., {f}_{sim}^{k}]}$. And then we compute the spatial attention cost volume,
\begin{equation}
  \setlength\abovedisplayskip{3pt}
  \setlength\belowdisplayskip{3pt}
\begin{aligned}
  \emph{cv}_{spa} = \resizebox{0.36\hsize}{!}{$\frac{1}{\emph{k}}\sum\nolimits_{i=1}^k{{{f}_{sim}^{i} \cdot {f}_{t}^{s}}}$}.
  \label{eq:cv_attention}
\end{aligned}
\end{equation}

The fused cost volume ${\hat{\emph{cv}}_s}$ is produced by a residual operator as ${\hat{\emph{cv}}_s = \emph{cv}_{temp} + \alpha \emph{cv}_{spa}}$, where ${\alpha}$ denotes a fusion weight.
After that, we decode the fused cost volume to estimate optical flow ${\textbf{\emph{F}}_{syn}}$ of synthetic foggy images. We further transfer the pixel-wise motion knowledge from clean domain to synthetic foggy domain via flow consistency loss,
\begin{equation}
  \setlength\abovedisplayskip{3pt}
  \setlength\belowdisplayskip{3pt}
\begin{aligned}
\resizebox{0.13\hsize}{!}{$\mathcal{L}^{consis}_{flow}$} = \resizebox{0.28\hsize}{!}{$\sum\nolimits{||\textbf{\emph{F}}_{syn} - \textbf{\emph{F}}||_{1}}.$}
  \label{eqa:flow_consistency_loss}
\end{aligned}
\end{equation}
%

\subsection{Correlation-Alignment Motion Adaptation}
\begin{figure}
  \setlength{\abovecaptionskip}{5pt}
  \setlength{\belowcaptionskip}{-5pt}
  \centering
   \includegraphics[width=1.0\linewidth]{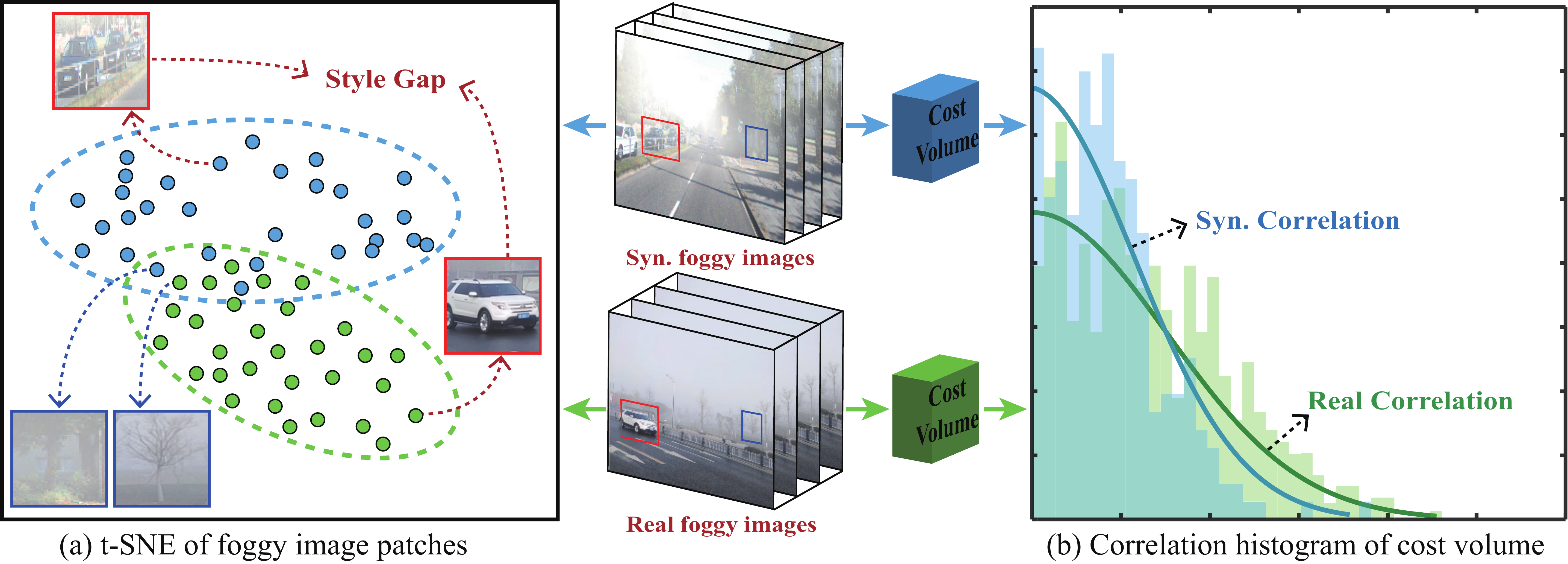}
   \caption{Visual distribution of synthetic and real foggy images. In (a) t-SNE of foggy image patches, the difference of degradation pattern is small, but there exists an obvious style gap between both the domains. In (b) correlation histogram of cost volume, the entire correlation distributions are similar. This motivates us to close the synthetic-to-real domain gap with correlation distribution.}
   \label{Correlation_Distribution}
\end{figure}

Although depth-association motion adaptation can bridge the clean-to-foggy domain gap and provide a coarse optical flow for synthetic foggy domain, it cannot help the synthetic-to-real domain gap. Hence, our method may inevitably suffer from artifacts under real foggy scenes due to the synthetic-to-real domain gap. To explore how large the synthetic-to-real foggy domain gap is, we visualize the feature distributions of the synthetic and real foggy images via t-SNE \cite{van2008visualizing} in Fig. \ref{Correlation_Distribution} (a). The degradation pattern discrepancy between synthetic and real foggy images is small, but there exists an obvious synthetic-to-real style gap that restricts the optical flow performance under real foggy scenes.

Direct motion adaptation from synthetic to real domain is difficult, since their background is different. Our solution is to construct an intermediate domain as an adaptation bridge namely cost volume, physically measuring the similarity between adjacent frames, not limited by scene difference.
We transform foggy images to cost volume space, and visualize the correlation distributions of synthetic and real foggy images via the histogram in Fig. \ref{Correlation_Distribution} (b). We can observe that both the domains share a similar correlation distribution. This motivates us to provide a novel correlation-alignment motion adaptation module, which can distill motion knowledge of the synthetic foggy domain to the real foggy domain.

\noindent
\textbf{Correlation Distribution Alignment (CDA).} We begin with two encoder ${\textbf{\emph{E}}_s}, {\textbf{\emph{E}}_r}$ for the synthetic foggy images ${[\textbf{\emph{J}}_{t}^{s}, \textbf{\emph{J}}_{t+1}^{s}]}$ and the real foggy images ${[\textbf{\emph{J}}_{t}^{r}, \textbf{\emph{J}}_{t+1}^{r}]}$, respectively.
We encode them to obtain the cost volume ${\emph{cv}_{s}, \emph{cv}_{r}}$ with the warp operator and the \textbf{SCA} module.
Furthermore, we randomly sample \emph{N} correlation values in the cost volumes ${\emph{cv}_{s}, \emph{cv}_{r}}$ to represent the entire correlation distribution of cost volume normalized into [0, 1] for both the domains. According to the range of correlation, we choose threshold values ${[\delta_1, \delta_2, ..., \delta_{k-1}]}$ to label the sampled correlation into \emph{k} classes, such as high correlation, and low correlation. Then, the correlation distribution \emph{p} is estimated by,
\begin{equation}
  \setlength\abovedisplayskip{3pt}
  \setlength\belowdisplayskip{3pt}
\begin{aligned}
  \emph{p} = \frac{\emph{n}+1}{N+\emph{k}},
  \label{eq:distribution}
\end{aligned}
\end{equation}
where \emph{n} is the number of the sampled correlation of one category. Note that we add an offset 1 to each category sampled correlation of Eq. \ref{eq:distribution} to ensure at least a single instance could be present in the real foggy domain. Thus, to align the features of the synthetic and real foggy domains, we minimize the correlation distribution distance between the two domains by enforcing Kullback-Leibler divergence,
\begin{equation}
  \setlength\abovedisplayskip{3pt}
  \setlength\belowdisplayskip{3pt}
\begin{aligned}
\resizebox{0.12\hsize}{!}{$\mathcal{L}^{kl}_{corr}$} = \resizebox{0.33\hsize}{!}{$\sum\nolimits_{i=1}^k{\emph{p}_{r,i}log\frac{\emph{p}_{r,i}}{\emph{p}_{s,i}}},$}
  \label{eqa:kl_div}
\end{aligned}
\end{equation}
where ${\emph{p}_{s,i}}$, ${\emph{p}_{r,i}}$ denote the \emph{i} category sampled correlation distributions of the synthetic foggy domain and the real foggy domain, respectively. The aligned correlation distributions represent that both the domains could have similar optical flow estimation capabilities. Finally, we decode the aligned cost volume to predict optical flow for real foggy images.

\noindent
\textbf{Self-Supervised Training Strategy.} To improve the robustness of knowledge transfer, we present a self-supervised training strategy that attempts to transfer motion knowledge from the synthetic foggy domain to the real foggy domain at the optical flow field level. We feed the real foggy images ${[\textbf{\emph{J}}_{t}^{r}, \textbf{\emph{J}}_{t+1}^{r}]}$ to the flow network of the synthetic foggy domain, which outputs the optical flow as the pseudo-labels ${\textbf{\emph{F}}_{pseudo}}$ (seeing the red arrow in Fig. \ref{Framework}). We then impose a self-supervised loss on the optical flow ${\textbf{\emph{F}}_{real}}$ estimated by the flow network of the real foggy domain,
\begin{equation}
  \setlength\abovedisplayskip{3pt}
  \setlength\belowdisplayskip{3pt}
\begin{aligned}
\resizebox{0.11\hsize}{!}{$\mathcal{L}^{self}_{flow}$} = \resizebox{0.38\hsize}{!}{$\sum\nolimits{||\textbf{\emph{F}}_{real} - \textbf{\emph{F}}_{pseudo}||_{1}}.$}
  \label{eqa:self_flow}
\end{aligned}
\end{equation}

During the training process, the encoder ${\textbf{\emph{E}}_{r}(f;{\theta}_{r})}$ of the real foggy domain is updated with the encoder ${\textbf{\emph{E}}_{s}(f;{\theta}_{s})}$ of the synthetic foggy domain using the exponential moving average (${\textbf{EMA}}$) mechanism,
namely, ${{\theta}_{r} \leftarrow {\theta}_{r} \cdot \lambda + {\theta}_{s} \cdot (1 - \lambda)}$, where ${\lambda}$ controls the window of ${\textbf{EMA}}$ and is often close to 1.0. The proposed correlation-alignment motion adaptation distills motion knowledge of the synthetic foggy domain to the real foggy domain in the feature correlation and optical flow dimensions, respectively.

\subsection{Total Loss and Implementation Details}
Consequently, the total objective for the proposed framework is written as follows,
\begin{equation}\footnotesize
  \begin{aligned}
  \setlength\abovedisplayskip{3pt}
  \setlength\belowdisplayskip{3pt}
  \mathcal{L} &= {\lambda}_1\mathcal{L}_{depth} + {\lambda}_2\mathcal{L}^{pho}_{flow} +  {\lambda}_3\mathcal{L}^{geo}_{flow} \\
   &+ {\lambda}_4\mathcal{L}^{consis}_{flow} +
  {\lambda}_5\mathcal{L}^{self}_{flow} + {\lambda}_6\mathcal{L}^{kl}_{corr},
  \label{total_loss}
\end{aligned}
\end{equation}
where the first four terms are the unsupervised losses that aim to transfer knowledge from the clean domain to the synthetic foggy domain, and the intention of the last two terms is to build the mathematical relationship between synthetic and real foggy domains. We empirically set the parameters $\{\lambda_1,\lambda_2,\lambda_3,\lambda_4,\lambda_5, \lambda_6\} = \{1,1,0.1,1,1,0.1\}$.
Besides, as for the parameters of \textbf{CDA}, we set the sample number \emph{N} as 1000 and the number of categories \emph{k} as 10. The classification threshold values ${\delta}$ are set linearly from [0, 1]. The weight ${\lambda}$ of the \textbf{EMA} for self-supervised training strategy is 0.99.

The proposed framework UCDA-Flow consists of three encoders, two flow decoders, one disp decoder, and one residual block for \textbf{SCA}. We first train the optical flow network and the disp network of the clean domain via $\mathcal{L}^{pho}_{flow}$, $\mathcal{L}_{depth}$ and $\mathcal{L}^{geo}_{flow}$.
We update the optical flow network of the synthetic foggy domain via $\mathcal{L}^{consis}_{flow}$. Then we update the optical flow network of the real foggy domain via $\mathcal{L}^{self}_{flow}$ and $\mathcal{L}^{kl}_{corr}$ with 1000 epochs and 0.0005 learning rate. After that, we optimize the whole framework via the full loss $\mathcal{L}$ with 0.0002 learning rate. At the test stage, the testing model only needs the optical flow network of the real foggy domain, including encoder, warp, cost volume, and flow decoder.

\begin{table*}\footnotesize
  \setlength{\abovecaptionskip}{3pt}
  \setlength\tabcolsep{3pt}
\centering
\caption{Quantitative results on synthetic Light Fog-KITTI2015 (LF-KITTI) and Dense Fog-KITTI2015 (DF-KITTI) datasets.}
\begin{tabular}{cc|cccccccccc}

\hline
\hline
\multicolumn{2}{c|}{\multirow{2}{*}{Method}}& \multicolumn{1}{c|}{\multirow{2}{*}{RobustFlow}} & \multicolumn{1}{c|}{\multirow{2}{*}{DenseFogFlow}} &
\multicolumn{3}{c|}{UFlow} & \multicolumn{3}{c|}{Selflow} & \multicolumn{1}{c|}{\multirow{2}{*}{SMURF}} & \multirow{2}{*}{UCDA-Flow}\\

\cline{5-7}\cline{8-10}
\multicolumn{2}{c|}{}& \multicolumn{1}{c|}{} & \multicolumn{1}{c|}{} & \multicolumn{1}{c|}{-} & \multicolumn{1}{c|}{FFA-Net +}& \multicolumn{1}{c|}{AECR-Net +}&  \multicolumn{1}{c|}{-}& \multicolumn{1}{c|}{FFA-Net +}& \multicolumn{1}{c|}{AECR-Net +}  & \multicolumn{1}{c|}{} &\\
 \hline
\multicolumn{1}{c|}{\multirow{2}{*}{LF-KITTI}} & EPE & \multicolumn{1}{c|}{23.48}& \multicolumn{1}{c|}{6.82} &\multicolumn{1}{c|}{14.33}& \multicolumn{1}{c|}{14.21}& \multicolumn{1}{c|}{11.66}& \multicolumn{1}{c|}{13.42}& \multicolumn{1}{c|}{13.15}& \multicolumn{1}{c|}{10.06}& \multicolumn{1}{c|}{10.48} & \textbf{5.94} \\
\cline{2-12}
\multicolumn{1}{c|}{}& F1-all & \multicolumn{1}{c|}{81.54\%} & \multicolumn{1}{c|}{39.18\%} & \multicolumn{1}{c|}{56.96\%}& \multicolumn{1}{c|}{56.38\%} & \multicolumn{1}{c|}{50.92\%} & \multicolumn{1}{c|}{55.37\%} & \multicolumn{1}{c|}{54.83\%} & \multicolumn{1}{c|}{48.74\%}  & \multicolumn{1}{c|}{47.60\%} & \textbf{34.11\%} \\
\hline
\multicolumn{1}{c|}{\multirow{2}{*}{DF-KITTI}} & EPE & \multicolumn{1}{c|}{25.32}& \multicolumn{1}{c|}{8.03} &\multicolumn{1}{c|}{16.55}& \multicolumn{1}{c|}{15.97}& \multicolumn{1}{c|}{12.16}& \multicolumn{1}{c|}{15.84}& \multicolumn{1}{c|}{14.93}& \multicolumn{1}{c|}{11.21}& \multicolumn{1}{c|}{11.56} & \textbf{6.29} \\
\cline{2-12}
\multicolumn{1}{c|}{}& F1-all & \multicolumn{1}{c|}{85.77\%} & \multicolumn{1}{c|}{41.73\%} & \multicolumn{1}{c|}{62.84\%}& \multicolumn{1}{c|}{61.69\%} & \multicolumn{1}{c|}{53.17\%} & \multicolumn{1}{c|}{58.81\%} & \multicolumn{1}{c|}{57.06\%} & \multicolumn{1}{c|}{50.25\%}  & \multicolumn{1}{c|}{51.39\%} & \textbf{36.25\%} \\
\hline
\hline
\end{tabular}
 \label{tab:quantitative_result}
\end{table*}

\begin{figure*}
  \setlength{\abovecaptionskip}{5pt}
  \setlength{\belowcaptionskip}{-7pt}
  \centering
   \includegraphics[width=0.99\linewidth]{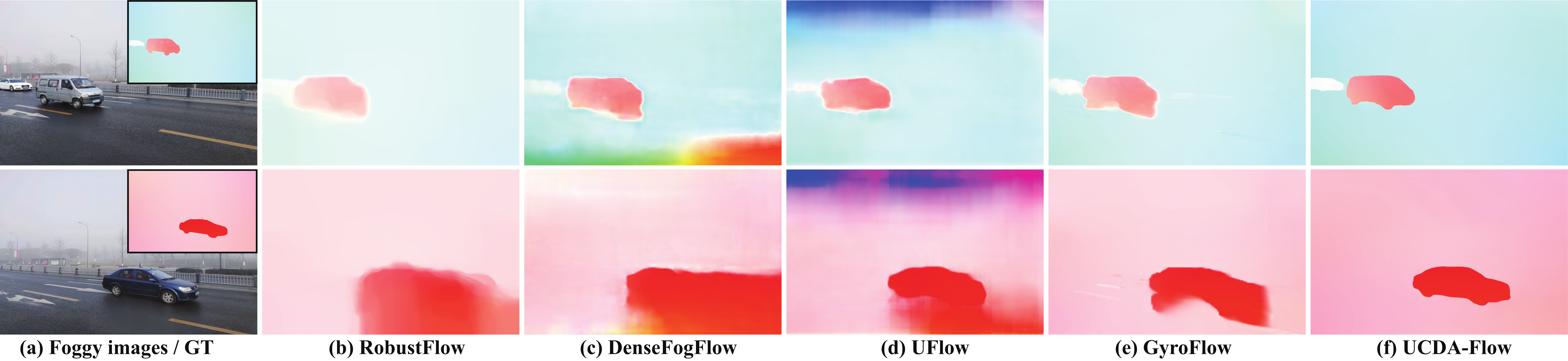}
   \caption{Visual comparison of optical flows on real Fog-GOF dataset.}
   \label{comparision_gof}
\end{figure*}

\begin{table}\footnotesize
  \setlength{\abovecaptionskip}{3pt}
  \setlength\tabcolsep{1pt}
\centering
\renewcommand\arraystretch{1.1}
\caption{Quantitative results on real foggy datasets.}
\begin{tabular}{cc|ccccccc}
\hline
\multicolumn{2}{c|}{\multirow{2}{*}{Method}}& \multicolumn{1}{c|}{\multirow{2}{*}{\makecell{Robust\\Flow}}} & \multicolumn{1}{c|}{\multirow{2}{*}{UFlow}} & \multicolumn{2}{c|}{\multirow{1}{*}{GMA}} & \multicolumn{1}{c|}{\multirow{2}{*}{\makecell{Dense\\FogFlow}}} & \multicolumn{1}{c|}{\multirow{2}{*}{\makecell{Gyro\\Flow}}} & \multicolumn{1}{c}{\multirow{2}{*}{Ours}} \\

\cline{5-6}
\multicolumn{2}{c|}{}& \multicolumn{1}{c|}{} & \multicolumn{1}{c|}{} & \multicolumn{1}{c|}{-} & \multicolumn{1}{c|}{ssl}& \multicolumn{1}{c|}{}&  \multicolumn{1}{c|}{}& \multicolumn{1}{c}{}\\
\hline
\multicolumn{1}{c|}{\multirow{2}{*}{\makecell{Fog-\\GOF}}}& \multicolumn{1}{c|}{EPE} & \multicolumn{1}{c|}{12.25} & \multicolumn{1}{c|}{2.97} & \multicolumn{1}{c|}{1.63}& \multicolumn{1}{c|}{1.69}&  \multicolumn{1}{c|}{1.78}& \multicolumn{1}{c|}{0.95} & \multicolumn{1}{c}{\textbf{0.81}}\\

\cline{2-9}
\multicolumn{1}{c|}{}& \multicolumn{1}{c|}{F1-all} & \multicolumn{1}{c|}{80.93\%} & \multicolumn{1}{c|}{30.82\%} & \multicolumn{1}{c|}{14.25\%}& \multicolumn{1}{c|}{15.11\%}&  \multicolumn{1}{c|}{16.41\%}& \multicolumn{1}{c|}{9.13\%} & \multicolumn{1}{c}{\textbf{7.18\%}}\\

\hline
\multicolumn{1}{c|}{\multirow{2}{*}{\makecell{Dense\\-Fog}}}& \multicolumn{1}{c|}{EPE} & \multicolumn{1}{c|}{13.48} & \multicolumn{1}{c|}{6.21} & \multicolumn{1}{c|}{3.68}& \multicolumn{1}{c|}{3.81}&  \multicolumn{1}{c|}{4.32}& \multicolumn{1}{c|}{-} & \multicolumn{1}{c}{\textbf{2.94}}\\

\cline{2-9}
\multicolumn{1}{c|}{}& \multicolumn{1}{c|}{F1-all} & \multicolumn{1}{c|}{79.31\%} & \multicolumn{1}{c|}{62.45\%} & \multicolumn{1}{c|}{33.18\%}& \multicolumn{1}{c|}{35.20\%}&  \multicolumn{1}{c|}{41.26\%}& \multicolumn{1}{c|}{-\%} & \multicolumn{1}{c}{\textbf{28.67\%}}\\
\hline
\end{tabular}
 \label{tab:gof_comparison}
\end{table}

\section{Experiments}
\subsection{Experiment Setup}
\noindent
\textbf{Dataset.}
We take the KITTI2015 \cite{menze2015object} dataset as the representative clean scene. We validate the performance of optical flow on one synthetic and three real foggy datasets.

\noindent
$\bullet$ \textbf{Fog-KITTI2015.}
We construct a synthetic foggy KITTI dataset with different densities of fog (\emph{e.g.}, dense fog and light fog) onto images of KITTI2015 \cite{menze2015object} via atmospheric scattering model \cite{narasimhan2002vision}. We select 8400 images of Fog-KITTI2015 dataset for training and 400 images for testing.

\noindent
$\bullet$ \textbf{Fog-GOF.} GOF \cite{li2021gyroflow} is a dataset containing four different scenes with synchronized gyro readings, such as regular scenes and adverse weather scenes. We choose foggy images of GOF to compose a new foggy dataset, namely Fog-GOF, of which 1000 images for training and 105 images for testing.

\noindent
$\bullet$ \textbf{DenseFog.} We seek the real foggy dataset collected by DenseFogFlow \cite{yan2020optical}, namely DenseFog, of which 2346 images for training and 100 images for testing.

\noindent
$\bullet$ \textbf{Real-Fog World.} We collect degraded videos under real foggy scenes from \emph{Youtube}, with 1200 and 240 images for training and testing, respectively.

\noindent
\textbf{Comparison Methods.}
We choose three competing methods GyroFlow \cite{li2021gyroflow}, DenseFogFlow \cite{yan2020optical} and RobustFlow \cite{li2018robust} which are designed for adverse weather optical flow. Moreover, we select several state-of-the-art supervised (GMA \cite{jiang2021transformer}) and unsupervised (SMURF \cite{stone2021smurf}, UFlow \cite{jonschkowski2020matters} and Selflow \cite{liu2019selflow}) optical flow approaches designed for clean scenes.
The unsupervised methods are first trained on KITTI2015 for initialization and re-trained on the target degraded dataset. The supervised method is first trained on the synthetic dataset, and then trained on target real datasets via self-supervised learning \cite{stone2021smurf}, denoted with `ssl' in Table. \ref{comparision_gof}. As for the comparison on Fog-KITTI2015, we design two different training strategies for competing methods. The first is that we directly train the comparison methods on foggy images. The second is to perform the defogging first via defog approaches (\emph{e.g.}, FFA-Net \cite{qin2020ffa} and AECR-Net \cite{wu2021contrastive}), and then we train the comparison methods on the defogging results (named as FFA-Net+ / AECR-Net+).

\noindent
\textbf{Evaluation Metrics.}
We choose average endpoint error (EPE \cite{dosovitskiy2015flownet}) and the lowest percentage of flow outliers (F1-all \cite{menze2015object}) as evaluation metrics for the quantitative evaluation. The smaller the index is, the better the predicted result is.

\begin{figure*}
  \setlength{\abovecaptionskip}{5pt}
  \setlength{\belowcaptionskip}{-12pt}
  \centering
   \includegraphics[width=0.99\linewidth]{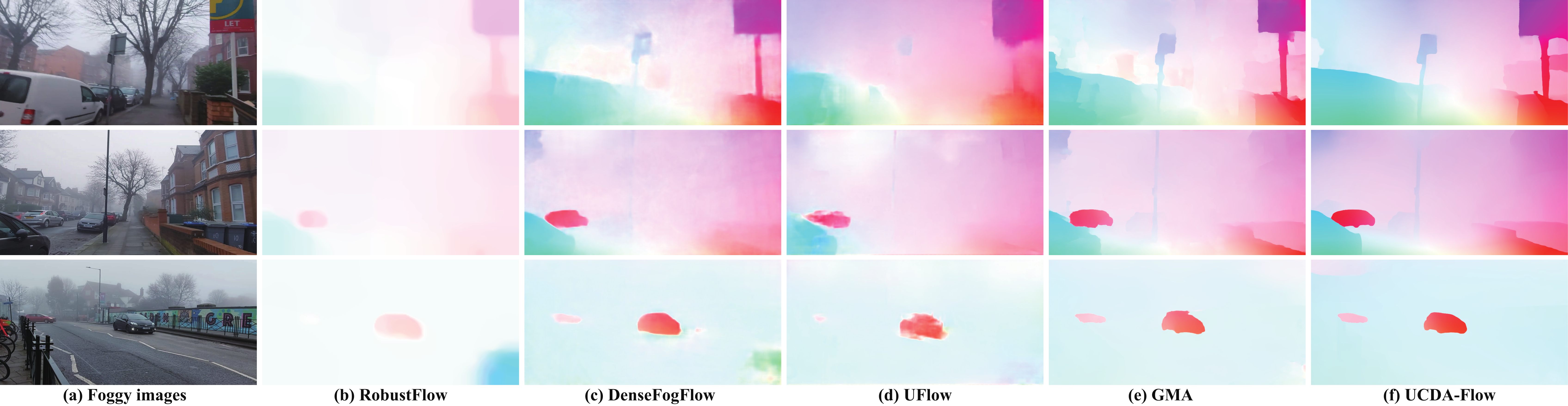}
   \caption{Visual comparison of optical flows on Real-Fog World.}
   \label{comparision_real}
\end{figure*}

\subsection{Experiments on Synthetic Images}
In Table \ref{tab:quantitative_result}, we show the quantitative comparison of the synthetic light and dense Fog-KITTI2015 datasets. Note that, we choose unsupervised methods for fair comparison which all do not need any ground truth.
We have two key observations. First, the proposed UCDA-Flow is significantly better than the unsupervised counterparts under light and dense foggy conditions. Since degradation breaks the basic assumption of optical flow, the competing methods cannot work well. Second, the pre-processing procedure defogging (\emph{e.g.}, FFA-Net / AECR-Net + UFlow) is positive to optical flow estimation. However, since the defog methods are not designed for optical flow and the defogging images may be over-smoothness, the performance of optical flow is still limited. On the contrary, the proposed method could well handle both light and dense foggy images. The reason is that the proposed UCDA-Flow bypasses the difficulties of directly estimating optical flow from degraded images, and transferring motion knowledge from clean domain to foggy domain via unsupervised domain adaptation.

\begin{figure}
  \setlength{\abovecaptionskip}{5pt}
  \setlength{\belowcaptionskip}{-5pt}
  \centering
   \includegraphics[width=1.0\linewidth]{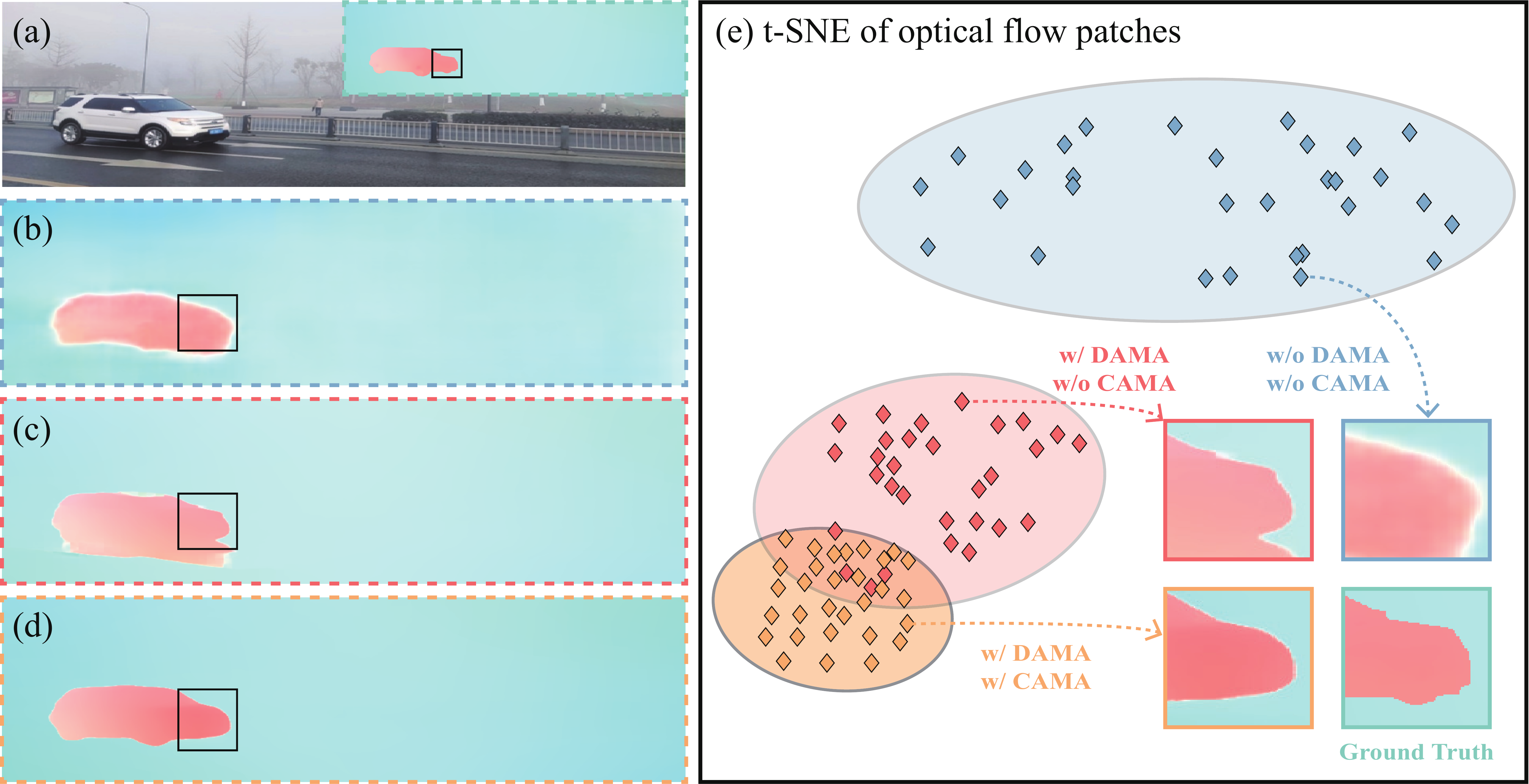}
   \caption{Effectiveness of cumulative adaptation architecture. (a) Real foggy image. (b)-(d) Optical flows without DA, with DAMA only, with DAMA and CAMA, respectively. (e) The t-SNE visualization of each adaptation strategy.}
   \label{correlation_adaptation}
\end{figure}

\subsection{Experiments on Real Images}
In Table \ref{tab:gof_comparison}, the quantitative results on Fog-GOF and DenseFog verify the superiority of our method. Note that, the performance barely changes before and after `ssl'. The reason is that fog has unexpectedly broken the photometric constancy assumption which self-supervised learning optical flow relies on, thus limiting its capability. In Fig. \ref{comparision_gof} and \ref{comparision_real}, we also show the visual comparison results on Fog-GOF and Real-Fog World datasets. The optimization-based method RobustFlow and the unsupervised method UFlow cannot work well. The supervised method GMA could obtain relatively smooth visualization results, but there exist outliers in Fig. \ref{comparision_real} (e). The hardware-assisted GyroFlow heavily relies on the camera ego-motion captured by the gyroscope data yet is less effective for the independent foreground object motion in Fig. \ref{comparision_gof} (e). DenseFogFlow only bridges the clean-to-foggy domain gap, but neglects the synthetic-to-real foggy domain gap, thus suffers artifacts when applied to real foggy images in Fig. \ref{comparision_gof} and \ref{comparision_real} (c).
On the contrary, the proposed cumulative adaptation framework can obtain satisfactory results under real foggy scenes in Fig. \ref{comparision_gof} and \ref{comparision_real} (f).

\begin{table}
\footnotesize
\setlength{\abovecaptionskip}{3pt}
\centering
\caption{Ablation study on adaptation losses.}
{
\begin{tabular}{cccc|cc}
      \hline
				\multirow{2}{*}{$\mathcal{L}^{consis}_{flow}$} & \multirow{2}{*}{$\mathcal{L}^{geo}_{flow}$} & \multirow{2}{*}{$\mathcal{L}^{self}_{flow}$} & \multirow{2}{*}{$\mathcal{L}^{kl}_{corr}$} & \multicolumn{2}{c}{Fog-GOF} \\
        \cline{5-6}
        & & & & EPE & F1-all \\
							 \hline
	         $\times$  &$\times$&$\times$& $\times$& 2.92 & 30.94\%  \\

          $\times$& $\surd$&$\times$&  $\times$& 2.88 & 30.20\% \\

			 $\surd$& $\times$&$\times$&  $\times$& 1.59 & 14.03\% \\

			$\surd$ 	&$\surd$ &$\times$&$\times$& 1.35 & 11.27\% \\

			$\surd$ &$\surd$ &$\surd$ & $\times$& 1.27 & 10.76\% \\
			$\surd$ 	&$\surd$ &$\times$&$\surd$  & 0.92 & 8.81\% \\
			$\surd$ 	&$\surd$ &$\surd$ &$\surd$ & \textbf{0.81} &\textbf{7.18\%} \\
         \hline
 \end{tabular}}
 \label{tab:losses}
\end{table}

\subsection{Ablation Study}
\noindent
\textbf{Effectiveness of Cumulative Adaptation Architecture.}
To illustrate the effectiveness of cumulative DAMA-CAMA architecture, in Fig. \ref{correlation_adaptation}, we show the optical flow estimation of different adaptation strategies and visualize their low-dimensional distributions via t-SNE. In Fig. \ref{correlation_adaptation} (b), we can observe that there exist artifacts in the motion boundary without domain adaptation. With DAMA only in Fig. \ref{correlation_adaptation} (c), most of the outliers caused by degradation are removed. with both DAMA and CAMA in Fig. \ref{correlation_adaptation} (d), the motion boundary is clearer. Moreover, we visualize their corresponding t-SNE distribution in Fig. \ref{correlation_adaptation} (e). The blue, red, and yellow diamonds denote the distributions without domain adaptation, with DAMA only and with DAMA-CAMA, respectively. The blue distribution is scattered, the red distribution is gradually focused, and the yellow distribution is most concentrated, illustrating that the cumulative domain adaptation could progressively improve real foggy scene optical flow.

\noindent
\textbf{Effectiveness of Adaptation Losses.}
We study how the adaptation losses of the proposed method contribute to the final result as shown in Table \ref{tab:losses}. $\mathcal{L}^{geo}_{flow}$ aim to enforce the optical flow in rigid regions. $\mathcal{L}^{consis}_{flow}$ is to transfer motion knowledge from the clean domain to the synthetic foggy domain. The goal of $\mathcal{L}^{self}_{flow}$ and $\mathcal{L}^{kl}_{corr}$ is to distill motion knowledge of the synthetic foggy domain to the real foggy domain. We can observe that the motion consistency loss $\mathcal{L}^{consis}_{flow}$ make a major contribution to the optical flow result, and the correlation distribution alignment loss $\mathcal{L}^{kl}_{corr}$ can further improve the optical flow under real foggy scenes.

\begin{table}
\footnotesize
\setlength{\abovecaptionskip}{3pt}
\setlength\tabcolsep{10pt}
\centering
   \renewcommand\arraystretch{1.1}
\caption{The effect of modules in CAMA stage on optical flow.}
{
\begin{tabular}{ccc|cc}
               \hline
				\multirow{2}{*}{\textbf{EMA}} & \multirow{2}{*}{\textbf{SCA}} & \multirow{2}{*}{\textbf{CDA}} & \multicolumn{2}{c}{Fog-GOF} \\
        \cline{4-5}
        & & & EPE & F1-all \\
							 \hline
	         $\times$  &$\times$&$\times$& 1.38 & 12.06\%  \\

			 $\surd$& $\times$&  $\times$& 1.36 & 11.43\% \\

			$\surd$ 	&$\surd$ &$\times$& 1.27 & 10.76\% \\

			$\surd$ 	&$\surd$ &$\surd$ & \textbf{0.81} &\textbf{7.18\%} \\
         \hline
 \end{tabular}}
 \label{tab:da_modules}
\end{table}

\begin{figure}
  \setlength{\abovecaptionskip}{5pt}
  \setlength{\belowcaptionskip}{-5pt}
  \centering
   \includegraphics[width=1.0\linewidth]{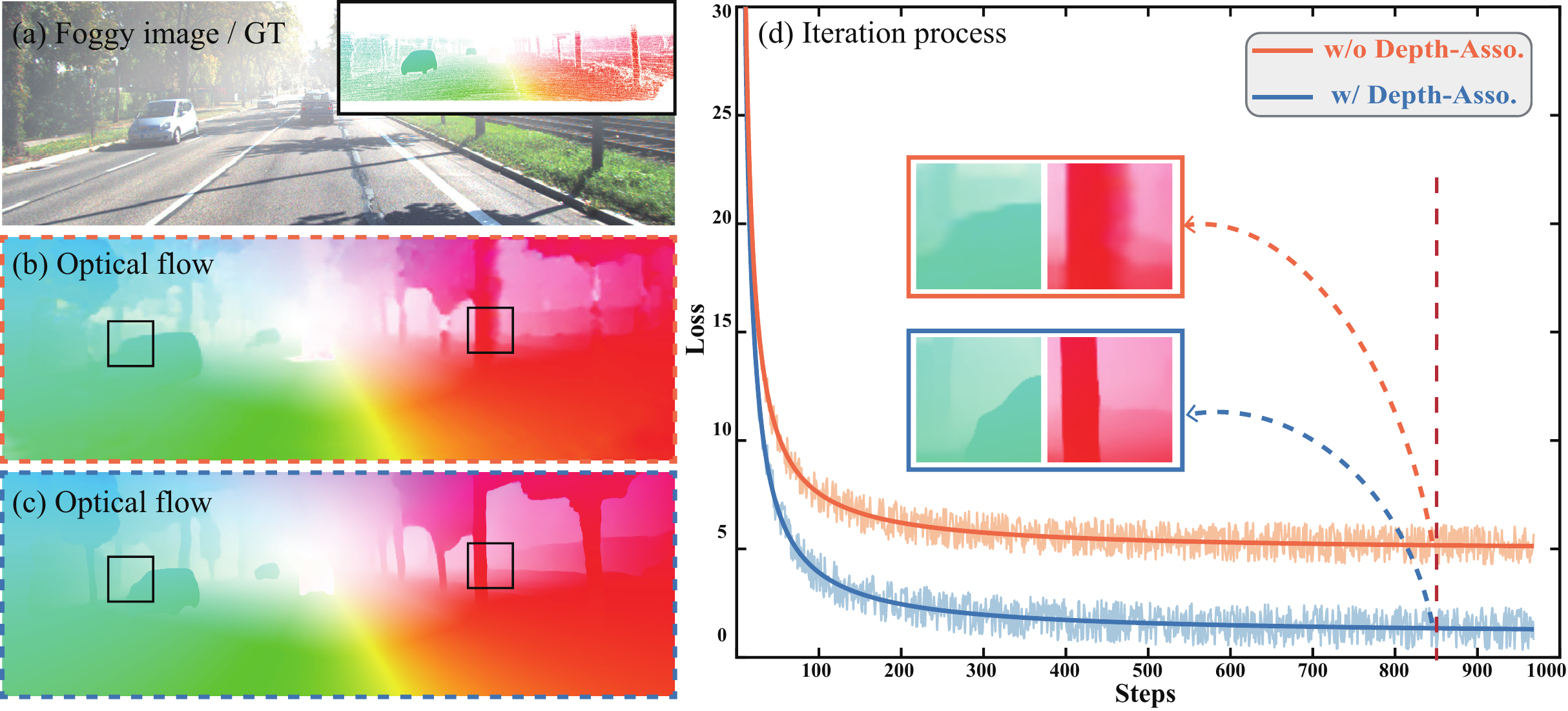}
   \caption{Effect of depth association on optical flow. (a) Foggy image. (b)-(c) Optical flows without depth-association and with depth-association, respectively. (d) Iteration process. Depth geometry association can improve the rigid motion boundary.
   }
   \label{Depth_Association}
\end{figure}

\subsection{Discussion}
\noindent
\textbf{How dose the Depth Improve Optical Flow?}
We study the importance of depth in transferring motion knowledge from clean domain to synthetic foggy domain in Fig. \ref{Depth_Association}.
Without depth association in Fig. \ref{Depth_Association} (b), the rigid motion boundaries are blurry. With depth association in Fig. \ref{Depth_Association} (c), the optical flow is global-smooth with sharp boundaries. Besides, we visualize their training iteration process in Fig. \ref{Depth_Association} (d). We can observe that depth association can further improve the optimal value that the proposed model converges to. Therefore, the depth geometry association can enhance the rigid motion boundary for synthetic foggy images.

\noindent
\textbf{Importance of Correlation Distribution Alignment.}
In Table \ref{tab:da_modules}, we show the effect of different modules on the optical flow of real foggy domain. \textbf{EMA} is to prevent the weights of the network from falling into the local optimum at the training stage. \textbf{SCA} aims to enhance the motion saliency in the cost volume.
\textbf{CDA} is to transfer motion knowledge from the synthetic foggy domain to the real foggy domain by aligning the correlation distributions of both the domains. As shown in Table \ref{tab:da_modules},  \textbf{EMA} and \textbf{SCA} contribute a small improvement on the optical flow, while the \textbf{CDA} plays a key role in improving the optical flow of real foggy domain.

\noindent
\textbf{Why Associate Depth with Fog?}
We also study the effect of different foggy image synthesis strategies on the optical flow in Table \ref{tab:foggy_generate_method}.
The GAN-based strategy cannot perform well. The reason is that GAN may erratically produce some new artifacts during the image translation, but instead exacerbate the synthetic-to-real foggy domain gap. On the contrary, since fog is a non-uniform degradation related to depth, it is reasonable that we use depth to synthesize foggy images. Note that, the depth estimated by monocular is not accurate enough due to the weak constraints.
We also upsample the sparse depth in KITTI dataset into the dense depth to synthesize the foggy images (pseudo-GT strategy), while this strategy is slightly inferior to our stereo-based strategy. The proposed depth association motion adaptation could make a positive contribution to motion knowledge transfer.

\begin{table}
\footnotesize
\setlength{\abovecaptionskip}{3pt}
\setlength\tabcolsep{8pt}
\centering
   \renewcommand\arraystretch{1.1}
\caption{Choice of different foggy image synthesis strategies.}
{
\begin{tabular}{cc|cc}
         \hline
				\multicolumn{2}{c|}{\multirow{2}{*}{Method}} & \multicolumn{2}{c}{Fog-GOF} \\
        \cline{3-4}
        & &  EPE & F1-all \\
							 \hline
	        \multicolumn{2}{c|}{GAN-Based}  &  1.43 & 13.10\%  \\
        \hline
        \multicolumn{1}{c|}{\multirow{3}{*}{Depth-Based}} &
			 Monocular  &  0.92 & 8.83\%  \\
       \cline{2-4}
       \multicolumn{1}{c|}{}&Pseudo-GT & 0.83 & 7.45\% \\
       \cline{2-4}
       \multicolumn{1}{c|}{}& \multicolumn{1}{c|}{\textbf{Stereo (Ours)}} &  \textbf{0.81} & \textbf{7.18\%} \\
         \hline
 \end{tabular}}
 \label{tab:foggy_generate_method}
\end{table}

\begin{figure}
  \setlength{\abovecaptionskip}{5pt}
  \setlength{\belowcaptionskip}{-5pt}
  \centering
   \includegraphics[width=1.0\linewidth]{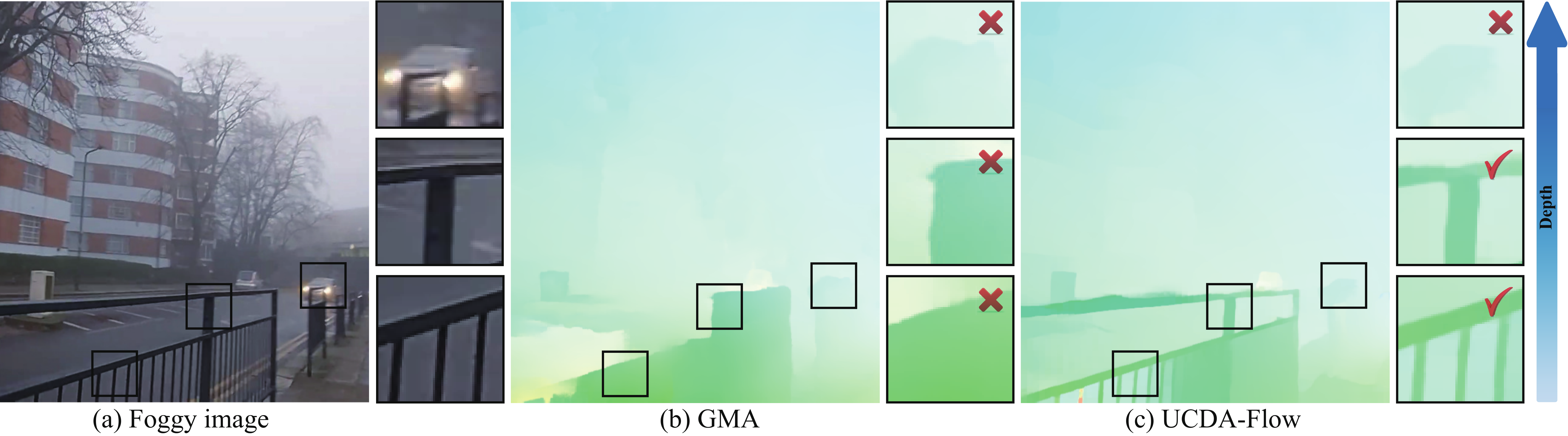}
   \caption{Limitation of the proposed method. Compared with the state-of-the-art optical flow method GMA \cite{jiang2021transformer}, UCDA-Flow obtains the clear motion boundary in the nearby regions, but fails for the too-distant moving objects under foggy scenes.}
   \label{limitation}
\end{figure}

\noindent
\textbf{Limitation.} In Fig. \ref{limitation}, we discuss the limitation of the proposed UCDA-Flow. Compared with the SOTA optical flow method GMA \cite{jiang2021transformer}, UCDA-Flow obtains the clearer motion boundary in the too-distantons, but fails for the too distant moving objects under foggy scenes. There are two reasons for this problem. First, our framework requires depth to enhance the optical flow in rigid regions, but it is difficult for the stereo strategy to obtain accurate depth in distant regions. Second, degradation is so severe that the details of the distant moving object are lost. In the future, we attempt to employ lidar for detecting distant objects.

\section{Conclusion}
In this work, we propose an unsupervised cumulative domain adaptation framework for optical flow under real foggy scenes. We reveal that depth is a key ingredient to influence optical flow, which motivates us to design a depth-association motion adaptation to close the clean-to-foggy domain gap. We figure out that cost volume correlation shares a similar distribution of the synthetic and real foggy images, which enlightens us to devise a correlation-alignment motion adaptation to bridge the synthetic-to-real domain gap.
We have conducted experiments on the synthetic and real foggy datasets to verify the superiority of our method.

\noindent
\textbf{Acknowledgments.} This work was supported in part by the National Natural Science Foundation of China under Grant 61971460, in part by JCJQ Program under Grant 2021-JCJQ-JJ-0060, in part by the National Natural Science Foundation of China under Grant 62101294, and in part by Xiaomi Young Talents Program.

{\small
\bibliographystyle{ieee_fullname}
\bibliography{egbib}

\begin{thebibliography}{10}\itemsep=-1pt

\bibitem{aleotti2021learning}
Filippo Aleotti, Matteo Poggi, and Stefano Mattoccia.
\newblock Learning optical flow from still images.
\newblock In {\em IEEE Conf. Comput. Vis. Pattern Recog.}, pages 15201--15211,
  2021.

\bibitem{chi2021feature}
Cheng Chi, Qingjie Wang, Tianyu Hao, Peng Guo, and Xin Yang.
\newblock Feature-level collaboration: Joint unsupervised learning of optical
  flow, stereo depth and camera motion.
\newblock In {\em IEEE Conf. Comput. Vis. Pattern Recog.}, pages 2463--2473,
  2021.

\bibitem{cho2014properties}
Kyunghyun Cho, Bart Van~Merri{\"e}nboer, Dzmitry Bahdanau, and Yoshua Bengio.
\newblock On the properties of neural machine translation: Encoder-decoder
  approaches.
\newblock {\em arXiv preprint arXiv:1409.1259}, 2014.

\bibitem{dosovitskiy2015flownet}
Alexey Dosovitskiy, Philipp Fischer, Eddy Ilg, Philip Hausser, Caner Hazirbas,
  Vladimir Golkov, Patrick Van Der~Smagt, Daniel Cremers, and Thomas Brox.
\newblock Flownet: Learning optical flow with convolutional networks.
\newblock In {\em Int. Conf. Comput. Vis.}, pages 2758--2766, 2015.

\bibitem{fu2017clearing}
Xueyang Fu, Jiabin Huang, Xinghao Ding, Yinghao Liao, and John Paisley.
\newblock Clearing the skies: A deep network architecture for single-image rain
  removal.
\newblock {\em IEEE Trans. Image Process.}, 26(6):2944--2956, 2017.

\bibitem{huang2022flowformer}
Zhaoyang Huang, Xiaoyu Shi, Chao Zhang, Qiang Wang, Ka~Chun Cheung, Hongwei
  Qin, Jifeng Dai, and Hongsheng Li.
\newblock Flowformer: A transformer architecture for optical flow.
\newblock In {\em Eur. Conf. Comput. Vis.}, pages 668--685. Springer, 2022.

\bibitem{hui2018liteflownet}
Tak-Wai Hui, Xiaoou Tang, and Chen~Change Loy.
\newblock Liteflownet: A lightweight convolutional neural network for optical
  flow estimation.
\newblock In {\em IEEE Conf. Comput. Vis. Pattern Recog.}, pages 8981--8989,
  2018.

\bibitem{ilg2017flownet}
Eddy Ilg, Nikolaus Mayer, Tonmoy Saikia, Margret Keuper, Alexey Dosovitskiy,
  and Thomas Brox.
\newblock Flownet 2.0: Evolution of optical flow estimation with deep networks.
\newblock In {\em IEEE Conf. Comput. Vis. Pattern Recog.}, pages 2462--2470,
  2017.

\bibitem{jiang2021transformer}
Shihao Jiang, Dylan Campbell, Yao Lu, Hongdong Li, and Richard Hartley.
\newblock Learning to estimate hidden motions with global motion aggregation.
\newblock In {\em Int. Conf. Comput. Vis.}, pages 9772--9781, 2021.

\bibitem{jiang2021learning}
Shihao Jiang, Yao Lu, Hongdong Li, and Richard Hartley.
\newblock Learning optical flow from a few matches.
\newblock In {\em IEEE Conf. Comput. Vis. Pattern Recog.}, pages 16592--16600,
  2021.

\bibitem{jonschkowski2020matters}
Rico Jonschkowski, Austin Stone, Jonathan~T Barron, Ariel Gordon, Kurt
  Konolige, and Anelia Angelova.
\newblock What matters in unsupervised optical flow.
\newblock In {\em Eur. Conf. Comput. Vis.}, pages 557--572. Springer, 2020.

\bibitem{kendall2015posenet}
Alex Kendall, Matthew Grimes, and Roberto Cipolla.
\newblock Posenet: A convolutional network for real-time 6-dof camera
  relocalization.
\newblock In {\em IEEE Conf. Comput. Vis. Pattern Recog.}, pages 2938--2946,
  2015.

\bibitem{lai2017semi}
Wei-Sheng Lai, Jia-Bin Huang, and Ming-Hsuan Yang.
\newblock Semi-supervised learning for optical flow with generative adversarial
  networks.
\newblock {\em Adv. Neural Inform. Process. Syst.}, 30, 2017.

\bibitem{li2021gyroflow}
Haipeng Li, Kunming Luo, and Shuaicheng Liu.
\newblock Gyroflow: Gyroscope-guided unsupervised optical flow learning.
\newblock pages 12869--12878, 2021.

\bibitem{li2018robust}
Ruoteng Li, Robby~T Tan, and Loong-Fah Cheong.
\newblock Robust optical flow in rainy scenes.
\newblock In {\em Eur. Conf. Comput. Vis.}, pages 288--304, 2018.

\bibitem{li2019rainflow}
Ruoteng Li, Robby~T Tan, Loong-Fah Cheong, Angelica~I Aviles-Rivero, Qingnan
  Fan, and Carola-Bibiane Schonlieb.
\newblock Rainflow: Optical flow under rain streaks and rain veiling effect.
\newblock In {\em Int. Conf. Comput. Vis.}, pages 7304--7313, 2019.

\bibitem{liu2020learning}
Liang Liu, Jiangning Zhang, Ruifei He, Yong Liu, Yabiao Wang, Ying Tai, Donghao
  Luo, Chengjie Wang, Jilin Li, and Feiyue Huang.
\newblock Learning by analogy: Reliable supervision from transformations for
  unsupervised optical flow estimation.
\newblock In {\em IEEE Conf. Comput. Vis. Pattern Recog.}, pages 6489--6498,
  2020.

\bibitem{liu2019ddflow}
Pengpeng Liu, Irwin King, Michael~R Lyu, and Jia Xu.
\newblock Ddflow: Learning optical flow with unlabeled data distillation.
\newblock In {\em AAAI Conf. on Arti. Intell.}, pages 8770--8777, 2019.

\bibitem{liu2019selflow}
Pengpeng Liu, Michael Lyu, Irwin King, and Jia Xu.
\newblock Selflow: Self-supervised learning of optical flow.
\newblock In {\em IEEE Conf. Comput. Vis. Pattern Recog.}, pages 4571--4580,
  2019.

\bibitem{liu2019griddehazenet}
Xiaohong Liu, Yongrui Ma, Zhihao Shi, and Jun Chen.
\newblock Griddehazenet: Attention-based multi-scale network for image
  dehazing.
\newblock In {\em Int. Conf. Comput. Vis.}, pages 7314--7323, 2019.

\bibitem{liu2021unpaired}
Yang Liu, Ziyu Yue, Jinshan Pan, and Zhixun Su.
\newblock Unpaired learning for deep image deraining with rain direction
  regularizer.
\newblock In {\em Int. Conf. Comput. Vis.}, pages 4753--4761, 2021.

\bibitem{luo2022learning}
Ao Luo, Fan Yang, Xin Li, and Shuaicheng Liu.
\newblock Learning optical flow with kernel patch attention.
\newblock In {\em IEEE Conf. Comput. Vis. Pattern Recog.}, pages 8906--8915,
  2022.

\bibitem{luo2021upflow}
Kunming Luo, Chuan Wang, Shuaicheng Liu, Haoqiang Fan, Jue Wang, and Jian Sun.
\newblock Upflow: Upsampling pyramid for unsupervised optical flow learning.
\newblock In {\em IEEE Conf. Comput. Vis. Pattern Recog.}, pages 1045--1054,
  2021.

\bibitem{meister2018unflow}
Simon Meister, Junhwa Hur, and Stefan Roth.
\newblock Unflow: Unsupervised learning of optical flow with a bidirectional
  census loss.
\newblock In {\em AAAI Conf. on Arti. Intell.}, pages 7251--7259, 2018.

\bibitem{menze2015object}
Moritz Menze and Andreas Geiger.
\newblock Object scene flow for autonomous vehicles.
\newblock In {\em IEEE Conf. Comput. Vis. Pattern Recog.}, pages 3061--3070,
  2015.

\bibitem{narasimhan2002vision}
Srinivasa~G Narasimhan and Shree~K Nayar.
\newblock Vision and the atmosphere.
\newblock {\em Int. J. Comput. Vis.}, 48(3):233--254, 2002.

\bibitem{qin2020ffa}
Xu Qin, Zhilin Wang, Yuanchao Bai, Xiaodong Xie, and Huizhu Jia.
\newblock Ffa-net: Feature fusion attention network for single image dehazing.
\newblock In {\em AAAI Conf. on Arti. Intell.}, pages 11908--11915, 2020.

\bibitem{ranjan2017optical}
Anurag Ranjan and Michael~J Black.
\newblock Optical flow estimation using a spatial pyramid network.
\newblock In {\em IEEE Conf. Comput. Vis. Pattern Recog.}, pages 4161--4170,
  2017.

\bibitem{ren2018gated}
Wenqi Ren, Lin Ma, Jiawei Zhang, Jinshan Pan, Xiaochun Cao, Wei Liu, and
  Ming-Hsuan Yang.
\newblock Gated fusion network for single image dehazing.
\newblock In {\em IEEE Conf. Comput. Vis. Pattern Recog.}, pages 3253--3261,
  2018.

\bibitem{ren2017unsupervised}
Zhe Ren, Junchi Yan, Bingbing Ni, Bin Liu, Xiaokang Yang, and Hongyuan Zha.
\newblock Unsupervised deep learning for optical flow estimation.
\newblock In {\em AAAI Conf. on Arti. Intell.}, pages 1495--1501, 2017.

\bibitem{shao2020domain}
Yuanjie Shao, Lerenhan Li, Wenqi Ren, Changxin Gao, and Nong Sang.
\newblock Domain adaptation for image dehazing.
\newblock In {\em IEEE Conf. Comput. Vis. Pattern Recog.}, pages 2808--2817,
  2020.

\bibitem{stone2021smurf}
Austin Stone, Daniel Maurer, Alper Ayvaci, Anelia Angelova, and Rico
  Jonschkowski.
\newblock Smurf: Self-teaching multi-frame unsupervised raft with full-image
  warping.
\newblock In {\em IEEE Conf. Comput. Vis. Pattern Recog.}, pages 3887--3896,
  2021.

\bibitem{sun2010secrets}
Deqing Sun, Stefan Roth, and Michael~J Black.
\newblock Secrets of optical flow estimation and their principles.
\newblock In {\em IEEE Conf. Comput. Vis. Pattern Recog.}, pages 2432--2439,
  2010.

\bibitem{sun2021autoflow}
Deqing Sun, Daniel Vlasic, Charles Herrmann, Varun Jampani, Michael Krainin,
  Huiwen Chang, Ramin Zabih, William~T Freeman, and Ce Liu.
\newblock Autoflow: Learning a better training set for optical flow.
\newblock In {\em IEEE Conf. Comput. Vis. Pattern Recog.}, pages 10093--10102,
  2021.

\bibitem{sun2018pwc}
Deqing Sun, Xiaodong Yang, Ming-Yu Liu, and Jan Kautz.
\newblock Pwc-net: Cnns for optical flow using pyramid, warping, and cost
  volume.
\newblock In {\em IEEE Conf. Comput. Vis. Pattern Recog.}, pages 8934--8943,
  2018.

\bibitem{teed2020raft}
Zachary Teed and Jia Deng.
\newblock Raft: Recurrent all-pairs field transforms for optical flow.
\newblock In {\em Eur. Conf. Comput. Vis.}, pages 402--419, 2020.

\bibitem{van2008visualizing}
Laurens Van~der Maaten and Geoffrey Hinton.
\newblock Visualizing data using t-sne.
\newblock 9(11), 2008.

\bibitem{wu2021contrastive}
Haiyan Wu, Yanyun Qu, Shaohui Lin, Jian Zhou, Ruizhi Qiao, Zhizhong Zhang, Yuan
  Xie, and Lizhuang Ma.
\newblock Contrastive learning for compact single image dehazing.
\newblock pages 10551--10560, 2021.

\bibitem{xu2020aanet}
Haofei Xu and Juyong Zhang.
\newblock Aanet: Adaptive aggregation network for efficient stereo matching.
\newblock In {\em IEEE Conf. Comput. Vis. Pattern Recog.}, pages 1959--1968,
  2020.

\bibitem{yan2020optical}
Wending Yan, Aashish Sharma, and Robby~T Tan.
\newblock Optical flow in dense foggy scenes using semi-supervised learning.
\newblock In {\em IEEE Conf. Comput. Vis. Pattern Recog.}, pages 13259--13268,
  2020.

\bibitem{yan2021self}
Wending Yan, Robby~T Tan, Wenhan Yang, and Dengxin Dai.
\newblock Self-aligned video deraining with transmission-depth consistency.
\newblock In {\em IEEE Conf. Comput. Vis. Pattern Recog.}, pages 11966--11976,
  2021.

\bibitem{yang2019volumetric}
Gengshan Yang and Deva Ramanan.
\newblock Volumetric correspondence networks for optical flow.
\newblock {\em Adv. Neural Inform. Process. Syst.}, 32:794--805, 2019.

\bibitem{yang2017deep}
Wenhan Yang, Robby~T Tan, Jiashi Feng, Jiaying Liu, Zongming Guo, and Shuicheng
  Yan.
\newblock Deep joint rain detection and removal from a single image.
\newblock In {\em IEEE Conf. Comput. Vis. Pattern Recog.}, pages 1357--1366,
  2017.

\bibitem{jason2016back}
Jason~J Yu, Adam~W Harley, and Konstantinos~G Derpanis.
\newblock Back to basics: Unsupervised learning of optical flow via brightness
  constancy and motion smoothness.
\newblock In {\em Eur. Conf. Comput. Vis.}, pages 3--10, 2016.

\bibitem{zhang2021separable}
Feihu Zhang, Oliver~J Woodford, Victor~Adrian Prisacariu, and Philip~HS Torr.
\newblock Separable flow: Learning motion cost volumes for optical flow
  estimation.
\newblock In {\em Int. Conf. Comput. Vis.}, pages 10807--10817, 2021.

\bibitem{zhang2018density}
He Zhang and Vishal~M Patel.
\newblock Density-aware single image de-raining using a multi-stream dense
  network.
\newblock In {\em IEEE Conf. Comput. Vis. Pattern Recog.}, pages 695--704,
  2018.

\bibitem{zhao2020maskflownet}
Shengyu Zhao, Yilun Sheng, Yue Dong, Eric~I Chang, Yan Xu, et~al.
\newblock Maskflownet: Asymmetric feature matching with learnable occlusion
  mask.
\newblock In {\em IEEE Conf. Comput. Vis. Pattern Recog.}, pages 6278--6287,
  2020.

\bibitem{zhou2017unsupervised}
Tinghui Zhou, Matthew Brown, Noah Snavely, and David~G Lowe.
\newblock Unsupervised learning of depth and ego-motion from video.
\newblock In {\em IEEE Conf. Comput. Vis. Pattern Recog.}, pages 1851--1858,
  2017.

\bibitem{zou2018df}
Yuliang Zou, Zelun Luo, and Jia-Bin Huang.
\newblock Df-net: Unsupervised joint learning of depth and flow using
  cross-task consistency.
\newblock In {\em Eur. Conf. Comput. Vis.}, pages 1--18. Springer, 2018.

\end{thebibliography}
}

\end{document}